\documentclass{article}

\usepackage{PRIMEarxiv}

\usepackage[utf8]{inputenc} 
\usepackage[T1]{fontenc}    
\usepackage{hyperref}       
\usepackage{url}            
\usepackage{booktabs}       
\usepackage{amsfonts}       
\usepackage{nicefrac}       
\usepackage{microtype}      
\usepackage{lipsum}
\usepackage{fancyhdr}       
\usepackage{graphicx}       
\graphicspath{{media/}}     
\usepackage[table,xcdraw]{xcolor}
\usepackage{subfigure}

\title{Time Series Representation Models}

\author{%
Robert Leppich$^{1*}$ \quad Vanessa Borst$^{1}$ \quad Veronika Lesch$^2$ \quad Samuel Kounev$^1$  \\
$^1$University of Wuerzburg, DE, \texttt{\{firstname.surname@uni-wuerzburg.de\}} \\
$^2$Cooperative State University Mosbach, DE, \texttt{veronika.lesch@dhbw.de}
}

\newcommand{\EncodingLayer}{\textit{EL}}
\newcommand{\RepresentationLayer}{\textit{RL}}
\newcommand{\MergeLayer}{\textit{ML}}
\newcommand{\AttMapClassifier}{\textit{AC}}

\begin{document}
\maketitle

\begin{abstract}
Time series analysis remains a major challenge due to its sparse characteristics, high dimensionality, and inconsistent data quality.
Recent advancements in transformer-based techniques have enhanced capabilities in forecasting and imputation; however, these methods are still resource-heavy, lack adaptability, and face difficulties in integrating both local and global attributes of time series.
To tackle these challenges, we propose a new architectural concept for time series analysis based on introspection. Central to this concept is the self-supervised pretraining of Time Series Representation Models (TSRMs), which once learned can be easily tailored and fine-tuned for specific tasks, such as forecasting and imputation, in an automated and resource-efficient manner.
Our architecture is equipped with a flexible and hierarchical representation learning process, which is robust against missing data and outliers. It can capture and learn both local and global features of the structure, semantics, and crucial patterns of a given time series category, such as heart rate data.
Our learned time series representation models can be efficiently adapted to a specific task, such as forecasting or imputation, without manual intervention. Furthermore, our architecture's design supports explainability by highlighting the significance of each input value for the task at hand.
Our empirical study using four benchmark datasets shows that, compared to investigated state-of-the-art baseline methods, our architecture improves imputation and forecasting errors by up to 90.34\% and 71.54\%, respectively, while reducing the required trainable parameters by up to 92.43\%. The source code is available at \url{https://github.com/RobertLeppich/TSRM}.
\end{abstract}


\section{Introduction}

%
%
%
%

Time series analysis yields high potential in both science and industry. It comprises various disciplines including time series forecasting, classification, and imputation. By analyzing time series data, we can gain deeper insights into various systems, such as sensor networks~\cite{papadimitriou2006optimal}, finance~\cite{zhu2002statstream}, and biological systems like the human body~\cite{ek2023transformer}. 

However, time series (TS) often exhibit high dimensionality, where each data point's relationship to other data points is determined by its temporal and attribute-level order. Most current approaches rely on prepossessing methods like seasonal-trend decomposition to address this complexity. Nevertheless, these methods usually depend on specific TS properties, such as stationarity, which may only sometimes be present or achievable~\cite{cleveland1990stl}. Furthermore, TS data from natural sources, such as electrocardiogram (ECG) readings or heart rate measurements during physical activity, present modeling challenges due to non-stationarity, dynamic patterns, noise, or missing data. 

Despite advancements in methodologies to tackle those challenges, like Box-Jenkins, Recurrent Neural Networks (RNNs), and Convolutional Neural Networks (CNNs), high dimensionality and vanishing/exploding gradients persist, restricting information flow over long sequences, as observed by Hochreiter et al.~\cite{hochreiter2001gradient}.
With the work of Vaswani et al.~\cite{vaswani2017attention}, the Transformer architecture was proposed and soon adapted into the domain of TS analysis~\cite{wu2020deep}. However, due to its design for the domain of Natural Language Processing (NLP) and the resulting use of the point-wise attention mechanism on word embeddings, approaches based on this architecture could not adequately capture all relevant TS characteristics and suffered from high computational and memory demands with long-term sequences~\cite{huang2018improved, povey2018time}. With time, more specialized implementations emerged, focusing on different subdomains of time-series analysis.  
TS forecasting started with improvements to the basic Transformer architecture to overcome the memory bottleneck~\cite{li2019enhancing}. Afterward, various enhancements, especially to the attention part, were quite successful, such as Informer by Zhou et al~\cite{zhou2021informer}, Autoformer by Wu et al.~\cite{wu2021autoformer}, and FEDFormer by Zhour et al.~\cite{zhou2022fedformer}. 
Transformer-based TS imputation evolved around CNN, RNN, auto-encoder, and GAN concepts~\cite{cao2018brits, fortuin2020gp, luo2018multivariate}, with recent successes such as SAITS from Du et al.~\cite{du2023saits}, which uses a joint multi-head self-attention approach for efficient imputation of missing values. 

However, despite all successes across most domains, current TS analysis approaches are still resource-heavy, lack adaptability, and face difficulties in integrating both local and global attributes of TS. Furthermore, most architectures are developed from scratch and tuned precisely to the downstream task. Such specialized models are often overoptimized and not very versatile when applied to other tasks~\cite{hendrycks2019using}. 

To overcome the challenges and limitations described above, in this paper we present a new architectural concept to tackle the complexity of TS analysis using time series representation models (TSRMs). TSRMs can be likened to language models in the field of Natural Language Processing (NLP), akin to Devlin et al.'s BERT~\cite{devlin2018bert}, as they act as a trained model that acquires an understanding of semantics, syntax, and critical attributes within a specific category of TS, such as electricity consumption, traffic volume, or electrocardiograms. 
Similar to the linguistic rules governing human language, analogous rules can be established for certain categories of TS data. Consider a univariate TS representing an athlete's heart rate.
While the heart rate dynamics reflect the athlete's current activities and are complex, they follow systematic rules and semantics such as a larger amplitudes during exertion compared to relaxation periods.

The method we propose consists of two phases: Pretraining and fine-tuning. Pretrained models demonstrate improved robustness and uncertainty estimation capabilities compared to models developed from scratch~\cite{hendrycks2019using}. In this work, pretraining is about learning and capturing the underlying patterns and specifics of TS categories, such as heart rate, within a pretrained TSRM. The pretraining phase is designed to be self-supervised, resource-efficient, and robust to outliers and missing data. 
After this pretraining phase, the TSRM can be adapted to specific tasks, such as TS forecasting, in a fine-tuning phase. During fine-tuning, the model efficiently uses the learned representations of the trained TSRM to optimize the downstream task, e.g. imputation, while maintaining its resilience to outliers and missing values. 
We summarize the key contributions of this work as follows:

\begin{itemize}
    \item We introduce Time Series Representation Models (TSRM) as novel architectural concept for TS analysis. 
    \item We propose the encoding layer (\EncodingLayer{}), designed to efficiently extract, enhance, and encode representations from time-series data while maintaining its dimensional integrity and providing robustness against outlier and missing values.
    \item We suggest three self-supervised pretraining tasks aimed at compelling the model to assimilate patterns, semantics, and distinct characteristics inherent in a TS category.
    \item We conduct extensive experiments across two temporal sequence tasks (imputation and forecasting), utilizing five benchmark datasets. 
    \item We investigate and discuss the architectures capabilities with respect to explainability.
\end{itemize}


\section{Methodology}\label{sec:methodology}

Our research presents a methodology that entails the initial pretraining of a temporal data model explicitly catered to a certain category of time series (TS), such as electrocardiogram (ECG). The pretrained model can then be fine-tuned for a specific task, such as TS forecasting. Consequently, performing model training from scratch for each task is unnecessary. Instead, individual training is only required for each category of TS. We utilize the terminology \textit{time~series~category} to classify a collection of TS that share fundamental rules and semantics analogous to diverse languages like German or English. Our methodological concept can be summarized as follows:

\begin{itemize}
    \item Pretraining on a TS category resulting in a TSRM that can effectively be fine-tuned for different downstream tasks.
    \item The pretraining is structured to facilitate self-supervision and maintain robustness against missing data. It comprises three artificially generated tasks to cultivate representations for a designated TS category efficiently. See Section~\ref{sec:meth_pretrain} for details.
    \item The fine-tuning process employs a pretrained TSRM to adapt effectively for a specific downstream task. The architecture autonomously adjusts to the task while maintaining resilience against missing data. See Section~\ref{meth:sec_fine_tuning} for details.
    \item We introduce a novel architecture for TS analysis, incorporating efficient layers designed to autonomously learn, enhance, and aggregate key patterns and details that are characteristic for a specific TS category. See Section~\ref{section:model_architecture} for details.
\end{itemize}

\subsection{Model Architecture}\label{section:model_architecture}
\begin{figure*}[]
\centering
         \centering
        \includegraphics[width=0.8\textwidth]{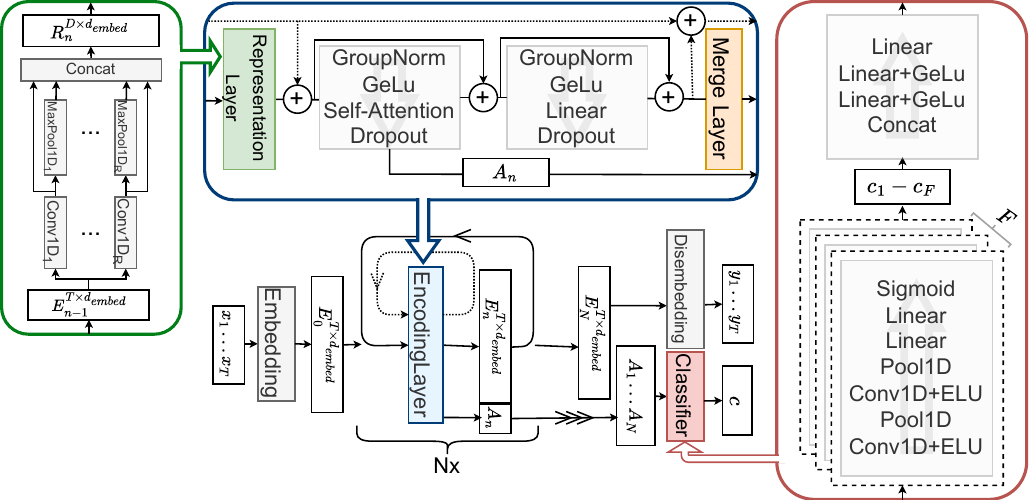}

\caption{Illustration of the proposed Time Series Representation Models (TSRM) framework, primarily composed of $N$ encoding layers (\EncodingLayer{}s) (upper section in blue), accompanied by the representation layer (\RepresentationLayer{}) (left, in green), and the attention map classifier (\AttMapClassifier{}) (right, in red).}
\label{figure:app:architecture_all}
\end{figure*}

Our proposed architecture (see Figure~\ref{figure:app:architecture_all}) was inspired by the concept of the BERT architecture~\cite{devlin2018bert} from the NLP domain. 
The input sequence $x_{1} \dots x_{T}$, where $x_i \in \mathbb{R}^F$ and $F$ is the amount of input features, undergoes a position-wise operation transferring the input features $f \in F$ into the dimension $d_{embed}$, resulting in the embedded matrix $E_0^{T \times d_{embed}}$. Here, an individual weight matrix $w_f^{1 \times f_{embed}}$ is employed per feature to achieve separate feature encodings of dimension $f_{embed}$, which are then concatenated together as follows: $E_{f_1}^{T \times f_{embed}} \oplus E_{f_2}^{T \times f_{embed}} \oplus \dots \oplus E_{f_F}^{T \times f_{embed}}$. This results in the total embedding dimension of $F \times f_{embed} = d_{embed}$. Subsequently, the matrix $E_0^{T \times d_{embed}}$ undergoes processing via $N$ consecutive encoding layers (\EncodingLayer{}s), each tasked with deriving representations from the matrix. These layers utilize the input matrix $E_{n-1}^{T \times d_{embed}}$ received from the preceding \EncodingLayer{}, enabling a hierarchic representation learning and resulting in an new encoding $E_{n}^{T \times d_{embed}}$. The output of each \EncodingLayer{} $E_{n}^{T \times d_{embed}}$ is fed into the next, as well as a residual connection, marked with dotted lines in Figure~\ref{figure:app:architecture_all}. The residual connection bridges the representation matrices across the \EncodingLayer{}s. This layer stacking and residual connections facilitate a structured feature extraction similar to deep CNN frameworks known from computer vision~\cite{he2016deep}. Following $N$ \EncodingLayer{}s, we split the final matrix $E_{N}^{T \times d_{embed}}$ up into $E_{f_1}^{T \times f_{embed}}, E_{f_2}^{T \times f_{embed}}, \dots, E_{f_F}^{T \times f_{embed}}$, and again utilize separate weight matrices $w_f^{f_{embed} \times 1}$ to generate the output sequence $y_{1} \dots y_{T}$, where $y_i \in \mathbb{R}^F$. 
Each \EncodingLayer{} results, besides the encoding and the residual connection, in associated attention weight matrices $A_n$.
The resulting attention maps $A_{1}\dots A_{N}$ from each \EncodingLayer{} are passed into the attention map classifier (\AttMapClassifier{}) (see Sect.~\ref{sec:meth:classifier}), which interprets the learned representations encoded in these attention matrices. This results in the final classification outcome $c$ both in pretraining and, if relevant, during downstream classification.

\paragraph{Encoding Layer}
Unlike information-rich word embeddings in NLP, which help models learn language patterns effectively~\cite{selva2021review}, the informational value of a single point in time is naturally lower, only gaining information content when combined across time steps or the feature dimension. TS also tend to be noisy, reducing the information value of individual time points.
To mitigate these challenges, we introduce a self-attention-based \EncodingLayer{} utilizing a potent representation methodology capable of efficiently encapsulating both local and global contextual information. The \EncodingLayer{}, presented in Figure~\ref{figure:app:architecture_all} (top), is structured as follows:
It starts with the representation layer (\RepresentationLayer{}), tasked with acquiring representations embedded within the sequence and concludes with the merge layer (\MergeLayer{}), which aggregates all acquired representations. The \EncodingLayer{}s are structured to maintain the dimensionality of the input at the output, allowing for $N$ to be modulated as an independent hyperparameter unrelated to the initial sequence length.
Below, we explain each of the three steps in the order of execution.

The representation layer (\RepresentationLayer{}), illustrated in Figure~\ref{figure:app:architecture_all} (left in green), comprises a setup of $M$ unique 1D CNN layers and $M$ corresponding 1D max-pooling layers. Designed to capture and assemble features from various abstraction levels, some of the $M$ CNN layers employ small kernels without dilation for identifying basic features, such as sequence details, while others use medium-sized kernels with minimal dilation for intermediate feature recognition, or large kernels with significant dilation for detecting comprehensive features like trends. The number of individual CNN layers~($M$), as well as the kernel size and dilation, are hyperparameter.
To reduce the impact of noise as well as outliers and enable a higher level of abstraction, we employ dilation in larger kernels. By default, all kernels are configured with a stride of 50\% of the kernel size and as depth-wise convolution.
The outputs of the $M$ CNN layers are inputted into respective max-pooling layers to refine the feature representations further. Subsequently, the outcomes from both the $M$ CNN and the $M$ max-pooling layers are concatenated, effectively transforming the input matrix $E_{n-1}^{T \times d_{embed}}$ into the representation matrix $R_{n}^{D \times d_{embed}}$, were $D$ corresponds to the total length of the concatenated matrices. 
This aggregation encapsulates the encoded features spanning varied abstraction levels.

The merge layer (\MergeLayer{}) is designed to reverse the dimensional alterations caused by the \RepresentationLayer{} and to aggregate the discovered representations. It comprises $M$ 1D transpose CNN layers utilizing transposed kernels to invert the transformations applied by the corresponding CNN layer of the \RepresentationLayer{}, thereby reinstating the original data dimensions.
Therefore, the matrix $R_{n}^{D \times d_{embed}}$ is segmented in contrast to the concatenation in the \RepresentationLayer{}. The segments corresponding to the pooling layers of the \RepresentationLayer{} are discarded.
Each segment is then processed through its respective transpose CNN layer, generating matrices that possess dimensions identical to those before concatenation. 
These $M$ matrices are then merged using a feed-forward network, resulting in a matrix that has the exact dimensions of the original input matrix for the \RepresentationLayer{}, $E_{n}^{T \times d_{embed}}$.

Situated between the \RepresentationLayer{} and the \MergeLayer{}, two blocks of layers facilitate the extraction and amplification of features from the \RepresentationLayer{}. Each is encapsulated within a residual skip connection following the preactivation design paradigm~\cite{he2016identity}. The initial block starts with a group normalization module comprising $F$ groups, enabling a feature-agnostic normalization approach~\cite{wu2018group}, which is succeeded by a GeLu activation function, a multi-head self-attention mechanism, and a dropout operation. The implementation of the multi-head self-attention spans conventional attention akin to the Transformer~\cite{vaswani2017attention}, sparse self-attention mimicking Wu et al.~\cite{wu2020adversarial}, and \textit{Propsparse attention} as suggested by Zhou et al.~\cite{zhou2021informer}. The choice of attention mechanism is adjustable via hyperparameters. In contrast to other methods, our technique preserves the uniqueness of input features by utilizing $F$ distinct attention modules. We split the input $R_{n}^{D \times d_{embed}}$ into its $F$ segments of size $R_{n}^{D \times f_{embed}}$, process them in individual attention modules, and subsequently concatenate the results during post-processing back into $R_{n}^{D \times d_{embed}}$. The attention maps $a_{1}^{\prime} \dots a_{F}^{\prime}$, where $a_i^{\prime} \in \mathbb{R}^{D \times D}$, undergo summation along their last dimension, yielding $a_1 \dots a_F$, where $a_i \in \mathbb{R}^D$. Each \EncodingLayer{} outputs a dedicated attention map per feature, where we abbreviate the $F$ maps per \EncodingLayer{} with $A_n$ in the following. The resultant matrices are emitted by the \EncodingLayer{} and subsequently analyzed by the \AttMapClassifier{}, explained below. 
The sequence then advances to the second block, initiating again with a group normalization layer, followed by a GeLu activation, a linear layer, and a dropout. The linear layer embedded within this block constitutes the sole fully connected component, interfacing all $d_{embedd}$ dimensions, thereby facilitating the learning of inter-feature correlations. The remainder of the architecture, except the \AttMapClassifier{}, maintains feature-separation.
Residual connections interlink all representations across \EncodingLayer{}s. These connections are added to the resultant matrix from the \RepresentationLayer{}. Subsequently, the attention and feature-correlation augmented matrix is fused with the residual connection prior to the introduction of the \MergeLayer{}, fostering information propagation independent of the \RepresentationLayer{} and the \MergeLayer{}.

\paragraph{Attention-map Classifier}\label{sec:meth:classifier}
The \AttMapClassifier{}, depicted in Figure~\ref{figure:app:architecture_all} (right, in red), exclusively utilize the attention maps generated by the \EncodingLayer{} as their input and have no access to the processed sequence itself. Thus, the \AttMapClassifier{} heavily relies on the \RepresentationLayer{}'s ability to learn accurate representations of the TS as well as the capacity of the multi-head attention layer to assign weights to these representations effectively.
Due to the feature-separated attention modules, each \EncodingLayer{} yields $F$ attention maps $a_1 \dots a_F$, where $a_i \in \mathbb{R}^{D}$. We abbreviate the total quantity of all $F$ attention maps per \EncodingLayer{} with $A_n$.
The $N$ attention maps belonging to each feature undergo individual processing via two 1D CNN layers, succeeded by an ELU activation, a max-pooling layer, two subsequent linear layers, and a final sigmoid activation for feature-specific classification outcomes. These feature-wise classifications are concatenated and utilized in an ensemble-based approach, incorporating three linear layers with two GeLu activations in between to achieve the final classification.

\subsection{Training Process}
Our training procedure involves two distinct phases: pretraining and fine-tuning. During pretraining, the model undergoes a self-supervised, multi-task training process aimed at capturing the underlying semantics of a specific TS type, independent of the intended downstream task. Subsequently, in the fine-tuning phase, the pretrained model is further trained on a specific task, such as TS forecasting.


\subsubsection{Pretraining a Time Series Model}\label{sec:meth_pretrain}

In order to facilitate the pretraining of a TS model, we employ a strategy known as dependent multi-task learning. All tasks are self-supervised and resilient to missing values within the TS. As a result, there is no need for additional labels to train the TS model, nor for the prior application of imputation algorithms. 
Our methodology is comparable to the pretraining approach used in BERT, which employs various language-related tasks to train a language model~\cite{devlin2018bert}. 
For our pretraining, we have devised three tasks specific to the TS domain to achieve similar outcomes: (i)~reconstruction of the input TS, (ii)~imputation, and (iii)~binary classification.
The idea behind the TS reconstruction and imputation task is centered on training the model to grasp the inherent semantics and structure of the TS, thereby enhancing its performance in both areas. Subsequently, the binary classification task builds upon the outputs of the reconstruction and imputation, categorizing the TS based on attention weights generated by the \EncodingLayer{}s. This paradigm trains the model to determine meaningful attention weights, forcing the \EncodingLayer{}s to generate distinct representations vital for successful classification.

\paragraph{Reconstruction} 
The first task is to reconstruct the input TS in the form of the model's output TS. It requires the model to preserve the structure of the TS throughout its various \EncodingLayer{}s, which encompass \RepresentationLayer{} and \MergeLayer{}s.
By doing so, the model is able to retain crucial information from the TS as it progresses through these encoding blocks.

\begin{figure}[h]
\includegraphics[width=0.8\textwidth]{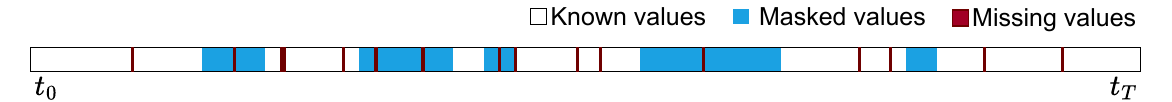}
\centering
\caption{Illustration of the artificial constructed imputation task during the pretraining step.}
\label{figure:app:imputation}
\end{figure}

\paragraph{Imputation}
For the artificially constructed imputation task depicted in Figure~\ref{figure:app:imputation}, contiguous time steps (marked blue) are selectively removed, and their original values are masked with $-1$, indicating a missing value. 
The masked subsets (with known values) serve as a basis for evaluating the effectiveness of the imputation task. 
The red sequences represent missing values in the TS that are also filled with $-1$ but will not be used for evaluation.
To prevent the model from simply learning the specific positions of missing values and to simulate a more realistic scenario, we evenly distribute the positions of the masked subsets throughout the TS, introduce a random shift to each subset, and randomly vary the size of each masked subset between $5\% - 10\%$ of the original input length. We set the amount of masked sequences to result in an overall masked subset of about $30\% - 50\%$ of the input TS. 


\paragraph{Binary Classification}
In addition to word masking, the developers of the BERT model implemented a next-sentence prediction (NSP) task to facilitate the acquisition of sentence relationships. 
In our study, we adapt this idea to the domain of TS analysis by a binary classification task requiring the model to categorize each TS as a viable candidate for the current TS category. 
Notably, we limit the classification task to depend solely on the reconstruction and imputation tasks by providing only the attention weights of the \EncodingLayer{}s to the \AttMapClassifier{} without providing access to the actual TS.
In cases involving univariate TS, we introduce artificial sequences with noise into the dataset, resulting in a binary classification task. In the case of multivariate TS, we rotate the feature space, which impacts $20\%$ of the dataset, implying that approximately every fifth TS is invalid and requires the model to classify it accordingly.

\paragraph{Loss calculation:} The loss function is composed of different components. The first part $L_{Repr.}$ measures the model's ability to reproduce the original TS output. This is evaluated using mean squared error (MSE) only for known, non-removed values, which are represented as white sequences in Figure~\ref{figure:app:imputation}. The second component $L_{Imp.}$ assesses the model's performance in imputing missing values using MSE for artificially removed values, which are represented as blue sequences in Figure~\ref{figure:app:imputation}. The third component $L_{Class.}$ evaluates the model's classification ability using binary cross entropy loss (BCE)~\cite{bce_pytorch}. The overall loss is calculated as the weighted sum of these components:

\begin{equation}
    Loss = (L_{Repr.} + L_{Imp.} \times L_\alpha) \times L_\beta + L_{Class.} \times L_\gamma 
\end{equation}{}

The scaling factors $L_\alpha$, $L_\beta$, and $L_\gamma$ are utilized to assign different weights to specific components of the pretraining process.
The exact weights for loss calculation will be provided for each experiment. Based on our preliminary experiments, we initially set $\alpha = 3.5$, $\beta = 1.2$, and $\gamma = 5$. As part of the classification task, the model is exposed to artificially generated TS data containing random noise or rotated feature spaces as explained above. In this scenario, we set $L_\alpha = L_\beta = 0$ to prevent the model from learning from invalid candidates of a TS category.

\subsubsection{Fine-tuning a time series model}\label{meth:sec_fine_tuning}
The process of fine-tuning utilizes a pretrained model to address specific TS tasks, such as forecasting. Fine-tuning is designed to be efficient and fast, requiring only minor modifications to the model's architecture, input data structure, and loss function. While our model is designed for a variety of tasks, including clustering and generating synthetic TS, in the following, we focus on the two most common tasks: prediction and imputation. We provide additional experiments for TS classification in the Appendix~\ref{appendix:meth_classification}.

During fine-tuning, we selectively disable gradient calculation of layers not associated with the task at hand. Specifically, gradients associated with the \AttMapClassifier{} are disabled during forecasting and imputation fine-tuning. Similarly, gradients relevant to the \EncodingLayer{} are disabled during classification operations.
This strategy reduces the count of trainable parameters, enhancing the procedural efficiency of fine-tuning.
Furthermore, we adapt the loss function to exclusively focus on the pertinent task. For instance, the loss associated with classification is deactivated during the fine-tuning phase for forecasting.

\paragraph{Forecasting and Imputation} For both tasks, we utilize the learned token ($-1$) to address missing values, given that the model knows this token from the pretraining phase and aims to replace it based on available data and learned features. 
Regarding imputation, the model has already handled this task during pretraining and allows adjustments during fine-tuning, where we again mark missing values with $-1$, which the model replaces with suitable values. 
In the prediction task, $-1$ tokens are used for both denoting the horizon and marking missing data in the input sequence. The model processes only this masked sequence and then predicts values corresponding to these tokens; only the values masked as the horizon are used to evaluate the effectiveness of the model. This process of replacing missing values with $-1$ tokens is further explained in Appendix~\ref{appendix:meth_forecasting}.

\section{Experiments}

In order to assess the effectiveness of our proposed architecture, we conducted a series of experiments using publicly accessible datasets from different fields: Electricity~\cite{dua2017uci} (univariate), Air-Quality~\cite{zhang2017cautionary} (multivariate), ETTm2~\cite{zhou2021informer} (univariate), and Traffic~\cite{traffic_dataset} (univariate). As Section~\ref{sec:methodology} outlines, the token $-1$ indicates absent data. Consequently, an adjustment of the used data is requisite, shifting all values into a positive range. A detailed description of the used datasets can be found in the Appendix~\ref{appendix:datasets}.

\subsection{Experimental Setup}\label{sec:exp_setup}
In each experiment, we compare our method with the state of the art (SOTA) in the respective field, following established benchmark configurations to increase comparability. 
Specifically, our methodology for imputation aligns with the framework outlined by Du et al.~\cite{du2023saits}, while for forecasting, we use the protocol established by Wu et al.~\cite{wu2021autoformer}. Both groups have each evaluated a number of base methods in their defined setting, which we also adopt in the subsequent comparisons.

For all experiments, we employed an early stopping with a threshold of an 1\% performance increase on the validation set with patience of 5 epochs. Hyper-parameter tuning was conducted through a systematic grid search, encompassing various parameters: count of stacked \EncodingLayer{}s $N$, number of attention heads $h$, encoding dimension $f_{embed}$, type of attention mechanism (vanilla, propsparse, sparse; see~\ref{section:model_architecture}), and configurations of the \RepresentationLayer{}, including the number of CNN layers, kernel dimensions, dilations, and grouping (see Appendix for details~\ref{appendix:exp_imputation1}.
Learning rates were initially determined with an automated range test from \textit{LightingAI}~(lightning.ai) and adapted during training with a learning rate scheduler from \textit{PyTorch} with a patience of two epochs. All models were trained with the Adam optimizer on an Nvidia A100 80GB GPU.

\subsection{Imputation}
To evaluate the performance of our architecture for TS imputation, we adopted the experimental setup of Du et al.~\cite{du2023saits} and introduced random data omissions into the Electricity dataset, resulting in two distinct missing rates: 10\% and 90\% (for 30\%, 50\%, 70\%, see Appendix~\ref{appendix:exp_imputation1}). In addition, we evaluate our model on the Air-Quality dataset using a missing rate of 10\%. We compare our results with reported results of SAITS~\cite{du2023saits}, BRITS~\cite{cao2018brits}, GP-VAE~\cite{fortuin2020gp}, M-RNN~\cite{yoon2018estimating}, and Transformer~\cite{vaswani2017attention}.

\subsection{Forecasting}
Regarding forecasting, we adopted the procedure of Wu et al.~\cite{wu2021autoformer}:
The input length is maintained at 96, while the prediction horizon is evaluated with 96 and 192, correspondingly. For the evaluation, we consider three datasets (i.e., Electricity, Traffic, and ETTm2) and compare against multiple SOTA techniques: FEDformer~\cite{zhou2022fedformer}, Autoformer~\cite{wu2021autoformer}, Informer~\cite{zhou2021informer}, LogTrans~\cite{li2019enhancing}, and Reformer~\cite{kitaev2020reformer}.

\section{Results}
We report our findings below, collecting all baseline results from the two approaches whose benchmark configurations we adopted respectively (i.e., from~\cite{du2023saits} for imputation and from~\cite{zhou2022fedformer} for forecasting).

\subsection{Imputation}\label{sec:meth:imputation}
We pretrained one TSRM for each dataset and fine-tuned them for imputation. We provide details about the employed TSRMs in the Appendix~\ref{appendix:exp_imputation1}.
Table~\ref{tab:imputation1} outlines the outcomes of our models, fine-tuned for an imputation task. For the Air-Quality dataset, we report the results for 10\% randomly omitted data and for Electricity for 10\% and 90\% randomly omitted data. We provide further results of additional setups in Appendix\ref{appendix:exp_imputation1}. Regarding the Electricity dataset with 10\% randomly omitted data, our model achieved a 90.34\% enhancement in performance using the MAE metric and a 62.72\% improvement utilizing the RMSE metric compared to the current most effective method, SAITS. The superior performance is also evident in the experiment with 90\% randomly omitted data, where the performance increases by 62.63\% compared to SAITS using the RMSE metric. Contrarily, within the Air-Quality dataset, our model demonstrated a 33.78\% advancement in the RMSE metric relative to SAITS, though it did not exhibit comparable progress in the MAE metric. We explore these variable performance metrics across datasets and include additional experimental results with the ETTm2 dataset in Appendix~\ref{appendix:exp_imputation1}.
Outcomes on the Electricity dataset at data removal rates of 30\% and 50\% for imputation demonstrate that our model proficiently performs the imputation task under substantial data deficiency.

Notably, our imputation models not only indicate efficiency but also achieve SOTA performance with a significantly reduced number of trainable parameters for both datasets, as documented in Table~\ref{tab:imputation1} in million~(M). 
For the Electricity dataset, our best model necessitated 0.86M trainable parameters, in contrast to SAITS, which required 11.51M. Similarly, for the Air-Quality dataset, our best model necessitated 0.98M trainable parameters, whereas SAITS needed 3.07M.

\begin{table}[]
\caption{Performance comparison for the imputation task.}
\centering
\label{tab:imputation1}
\scalebox{.78}{
\begin{tabular}{@{}llllllllll@{}}
\toprule
Dataset & \multicolumn{3}{c}{Air-Quality} &  & \multicolumn{5}{c}{Electricity} \\ \cmidrule(r){1-4} \cmidrule(l){6-10} 
Missing & \multicolumn{2}{c}{10\%} &  &  & \multicolumn{2}{c}{10\%} & \multicolumn{2}{c}{90\%} &  \\ \cmidrule(lr){2-3} \cmidrule(lr){6-9}
Metric & MAE & RMSE & P{[}M{]} &  & MAE & RMSE & MAE & RMSE $\downarrow$ & P{[}M{]} \\ \cmidrule(r){1-4} \cmidrule(l){6-10} 
M-RNN & 0.294 & 0.643 & 1.1 &  & 1.244 & 1.867 & 1.331 & 1.961 & 18.6 \\
BRITS & 0.153 & 0.525 & 11.3 &  & 0.847 & 1.322 & 1.163 & 1.702 & 7.00 \\
GP-VAE & 0.268 & 0.614 & \textbf{0.4} &  & 1.094 & 1.565 & 1.004 & 1.622 & 13.5 \\
Transf. & 0.158 & 0.521 & 5.1 &  & 0.823 & 1.301 & 0.934 & 1.492 & 14.8 \\
SAITS & \textbf{0.137} & 0.518 & 3.1 &  & 0.735 & 1.162 & 0.933 & 1.354 & 11.5 \\
TSRM & 0.157 & \textbf{0.343} & 1.0 &  & \textbf{0.071} & \textbf{0.146} & \textbf{0.273} & \textbf{0.506} & \textbf{0.86} \\ \bottomrule
\end{tabular}
}
\end{table}

\subsection{Forecasting}\label{sec:exp:forecasting}
We pretrained a TSRM for each dataset and chose the most effective variant, based on the performance during pretraining, for further optimization on the forecasting task. Additional information regarding these pretrained models is documented in the Appendix~\ref{appendix:exp_forecasting}. Outcomes for the optimized model on both the Electricity and Traffic datasets are outlined in Table~\ref{tab:forecasting1}. The sequence input was consistently held at a length of 96, with evaluation spans of 96 and 192. Notably, a single model was optimized for the longest evaluation span, utilizing truncation to accommodate the shorter span.

Results indicated a considerable enhancement across all evaluated metrics, including a boost in MSE performance by 71.28\% for the Electricity dataset and 99.42\% for the Traffic dataset over a horizon of 192. Although these results were remarkably positive, we could not beat the SOTA performance on the ETTm2 dataset. Further investigation revealed that the temporal embedding used by the listed baseline methods is a crucial component of a successful prediction, as these univariate input time series in particular do not provide enough information for a high-quality prediction. Despite experimenting with the same temporal embeddings, we could not improve the results for the ETTm2 dataset. This indicates that the architecture struggles to extract and utilize information from embeddings, such as temporal and positional, for its task. These issues are discussed further in Appendix~\ref{appendix:exp_forecasting}.

\begin{table}[]
\caption{Performance comparison for the forecasting task.}
\centering
\label{tab:forecasting1}
\scalebox{.78}{
\begin{tabular}{@{}lllllllllllllll@{}}
\toprule
 & \multicolumn{4}{c}{Electricity} &  & \multicolumn{4}{c}{Traffic} &  & \multicolumn{4}{c}{ETTm2} \\ \cmidrule(lr){2-5} \cmidrule(lr){7-10} \cmidrule(l){12-15} 
Horizon & \multicolumn{2}{c}{96} & \multicolumn{2}{c}{192} &  & \multicolumn{2}{c}{96} & \multicolumn{2}{c}{192} &  & \multicolumn{2}{c}{96} & \multicolumn{2}{c}{192} \\ \cmidrule(lr){2-5} \cmidrule(lr){7-10} \cmidrule(l){12-15} 
Metric & MSE$\downarrow$ & MAE & MSE & MAE &  & MSE & MAE & MSE & MAE &  & MSE & MAE & MSE & MAE \\ \cmidrule(lr){2-5} \cmidrule(lr){7-10} \cmidrule(l){12-15} 
Autoformer & 0.341 & 0.438 & 0.345 & 0.428 &  & 0.246 & 0.346 & 0.266 & 0.370 &  & \textbf{0.065} & \textbf{0.189} & 0.118 & 0.256 \\
LogTrans & 0.288 & 0.393 & 0.432 & 0.483 &  & 0.226 & 0.317 & 0.314 & 0.408 &  & 0.075 & 0.208 & 0.129 & 0.275 \\
Reformer & 0.275 & 0.379 & 0.304 & 0.402 &  & 0.313 & 0.383 & 0.386 & 0.453 &  & 0.077 & 0.214 & 0.138 & 0.290 \\
Informer & 0.258 & 0.367 & 0.285 & 0.388 &  & 0.257 & 0.353 & 0.299 & 0.376 &  & 0.080 & 0.217 & 0.112 & 0.259 \\
FEDformer & 0.253 & 0.370 & 0.282 & 0.386 &  & 0.170 & 0.263 & 0.173 & 0.265 &  & 0.072 & 0.206 & \textbf{0.102} & \textbf{0.245} \\
TSRM (Ours) & \textbf{0.072} & \textbf{0.114} & \textbf{0.081} & \textbf{0.123} &  & \textbf{0.001} & \textbf{0.018} & \textbf{0.001} & \textbf{0.018} &  & 0.072 & 0.196 & 0.128 & 0.260 \\ \bottomrule
\end{tabular}
}
\end{table}

\section{Explainability}\label{section:explain}
Our methodology's fundamental architectural principle is predicated on utilizing the attention mechanism as its central component and maintaining dimensional consistency throughout all \EncodingLayer{}s. 
As detailed in Section~\ref{sec:methodology}, the attention layers play a pivotal role in enhancing the representations from the representation layers.
This approach enables us to extract and investigate the attention weights of all \EncodingLayer{}s, offering valuable insights into our architecture's functioning and decision making. 
All $N$ \EncodingLayer{}s' attention layers output $F$ matrices, where $F$ reflects the number of features. This means that separate attention weights can be generated for each feature, thereby enabling their analysis in isolation. 
For this, 
it is imperative to revert the matrix dimensions dictated by the \RepresentationLayer{}s back to those of the input TS. This transformation employs the identical transpose CNN layer utilized in the \MergeLayer{}s, albeit with static weight matrices, designed to calculate the mean attention weight for each value. 
The $N$ back-transformed attention matrices can then be visualized together with the output to analyze the architecture's weighting during imputation, prediction, or classification. This can be done for all $N$ \EncodingLayer{}s individually or as a sum over all \EncodingLayer{}s to get an overview of all weights. 

\begin{figure*}
\includegraphics[width=\textwidth]{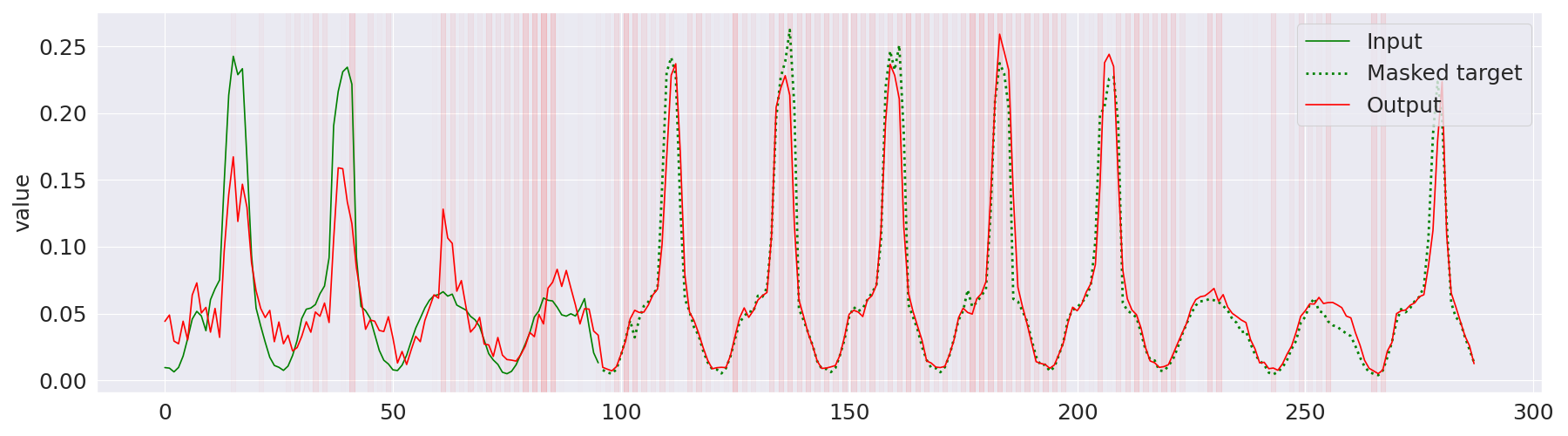}
\centering

\caption{Result of a fine-tuned forecasting model on the Traffic dataset with attention weights.}
\label{figure:exp:explain1}
\end{figure*}

Figure~\ref{figure:exp:explain1} shows an example with a fine-tuned forecasting model on the univariate Traffic dataset. The solid green line represents the initial input series, transitioning to a dotted line after the 96th value. This dotted green trajectory delineates the target horizon, integrated into the model as $-1$ values. The fine-tuned model appears to be less precise on the input sequence compared to the predicted sequence, showing the effect of the modified loss function during fine-tuning, which solely considers the forecasting horizon.
The vertical red bars are the reverted and summed attention weights from all $N$ \EncodingLayer{}s, indicating on what areas the attention was set on during the forecasting task. We observe the attention to focus around the last peeks before the prediction horizon. Further investigation revealed the model to set its attention only on the last peeks before the horizon at the first \EncodingLayer{} and shifted its attention to other areas in the following \EncodingLayer{}s. We provide a detailed interpretation of this example in Appendix~\ref{appendix:exp_explain1}.
\section{Limitations}\label{sec:limitations}
We evaluated our architecture by applying it to the benchmark datasets used in existing SOTA methods, including publicly accessible datasets from four different fields: Electricity~\cite{dua2017uci}, Air-Quality~\cite{zhang2017cautionary}, ETTm2~\cite{zhou2021informer}, and Traffic~\cite{traffic_dataset}. In the future, further datasets from other domains can be considered for a broader evaluation and validation of the effectiveness of our approach in scenarios beyond the ones examined in this paper. Moreover, examining contemporary approaches related to state space models~\cite{gu2022efficiently} and conducting an empirically robust comparison considering both performance and resource efficiency is crucial.
Additionally, it is necessary to develop an enhanced representation learning methodology. The existing system is promising due to its resource efficiency, as it maintains a constant number of trainable parameters independent of the dataset size. Nonetheless, the resultant representation matrix in the \RepresentationLayer{} $R_{n}^{D \times d_{embed}}$ expands with the dataset size, demanding corresponding memory resources. Consequently, this leads to slower computational processes when dealing with extensive input sequences. In addition, the optimization of knowledge retrieval, e.g. on the basis of time embedding, is still inadequate.
The explainability element of our proposed architecture demonstrates considerable potential for facilitating knowledge discovery in TS analysis. However, its efficacy has yet to be empirically evaluated using relevant benchmark datasets.

\section{Conclusion}
We proposed a novel architecture for TS analysis, which we refer to as Time Series Representation Model (TSRM). It employs a self-supervised pretraining mechanism to learn TSRMs specific to different categories of TS data. Following this, TSRMs allow effective adaptation for TS-specific applications, encompassing forecasting and imputation. Both stages of the training process for the TSRM are designed to maintain resilience against anomalies, such as missing data and outliers, while also offering a degree of explainability.
Empirical evaluations indicate that our proposed TSRM achieves enhanced performance and significantly reduces parameters compared to SOTA methods in the context of TS imputation and forecasting. We investigated current issues and potential for future work, especially regarding temporal embeddings and new fine-tuning tasks, such as TS clustering.

\bibliographystyle{unsrt}  
\bibliography{references}

\appendix
\section{Appendix}

\subsection{Datasets}\label{appendix:datasets}

\textbf{Electricity Load Diagram (Electricity)}: 
The Electricity dataset, available at UCI~\cite{dua2017uci}, contains electricity consumption data measured in kilowatt-hours (kWh). It includes data from 370 clients collected every 15 minutes for 48 months, starting from January 2011 to December 2014. 
Analog to baseline approaches, we divide the dataset into three parts: a test set (January 2011 to October 2011), a validation set (November 2011 to August 2012), and a training set (September 2012 to December 2014). To train our model, we select a sample every 100 consecutive steps. 

\textbf{Air-Quality}
The Air-Quality dataset~\cite{zhang2017cautionary} encompasses hourly measurements of atmospheric contaminants from 12 observational stations in Beijing. The collection spanned from 2013/03/01 to 2017/02/28, covering 48 months. Each station recorded 11 continuous variables (e.g., PM2.5, PM10, SO2), serving as features in our dataset. The dataset exhibits a missing data incidence of 1.6\%. The temporal division for experimental purposes classifies the first 10 months (2013/03 - 2013/12) as the test set, the subsequent 10 months (2014/01 - 2014/10) as the validation set, and the remaining 28 months (2014/11 - 2017/02) as the training set. Data samples were generated for our experiments by selecting every 24 consecutive hourly data points as a single sample. In a methodology akin to SAITS~\cite{du2023saits}, 10\% of observed values in both the validation and test sets were omitted to serve as ground-truth benchmarks for evaluation purposes.

\textbf{Traffic}
The Traffic dataset~\cite{traffic_dataset} comprehensively represents occupancy rates within California's freeway infrastructure. It is derived from 863 integrated sensor systems, which monitored vehicular capacity usage from July 2016 through 2018, culminating in 17,544 hourly records per sensor. This dataset is utilized as a univariate dataset for our analyses. The dataset was partitioned such that the initial 60\% constituted the training subset, the subsequent 20\% formed the validation subset, and the final 20\% comprised the test subset.



\textbf{Electricity Transformer Temperature (ETTm2)}:
The ETTm2 dataset comprises data collected from electricity transformers over a time period from July 1, 2016, to June 26, 2018. This dataset consists of samples collected every 15 minutes, resulting in 69,680 data points without any missing values. Each sample contains seven features, including oil temperature and six different types of external power load features~\cite{zhou2021informer}. The initial four months of data from July to October 2016 are designated as the test set. The subsequent four months from November 2016 to February 2017 are set aside as the validation set. The remaining sixteen months from March 2017 to June 2018 are utilized for training purposes. To generate time-series samples, a sliding-window method is employed with a window size of 6 hours (24 consecutive steps) and a sliding size of 3 hours (12 steps).

\textbf{Wireless Sensor Data Mining (WISDM)}: 
The data collection process involved 29 individuals carrying an android smartphone in their front pants pocket. These individuals performed six distinct activities, resulting in a dataset comprising 1,098,207 records. The data collected consisted of accelerometer readings at a frequency of 20Hz, with no instances of missing samples~\cite{kwapisz2011activity}. The finalized dataset includes the original time series of accelerometer data split into segments of 200 readings spanning 10 seconds. The dataset was partitioned such that the initial 60\% constituted the training subset, the subsequent 20\% formed the validation subset, and the final 20\% comprised the test subset.

\subsection{Experiments: Explainability}\label{appendix:exp_explain1}
\begin{figure}
\label{fig:appendix:explain}
     \centering
     \begin{subfigure}
         \centering
         \includegraphics[width=\textwidth]{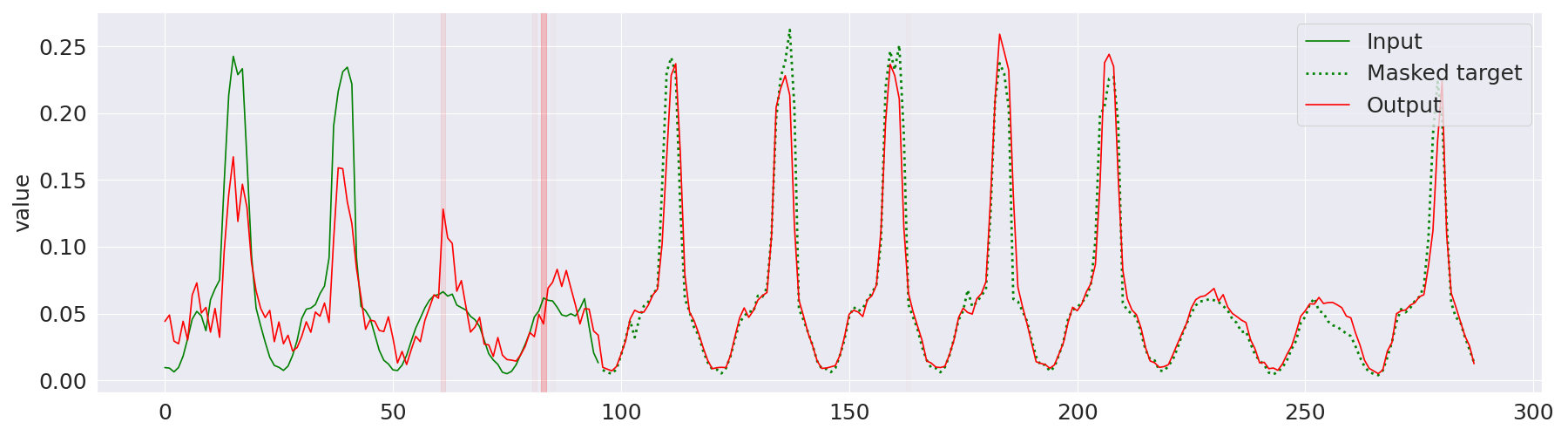}
     \end{subfigure} 
     \vfill 
     \begin{subfigure}
         \centering
         \includegraphics[width=\textwidth]{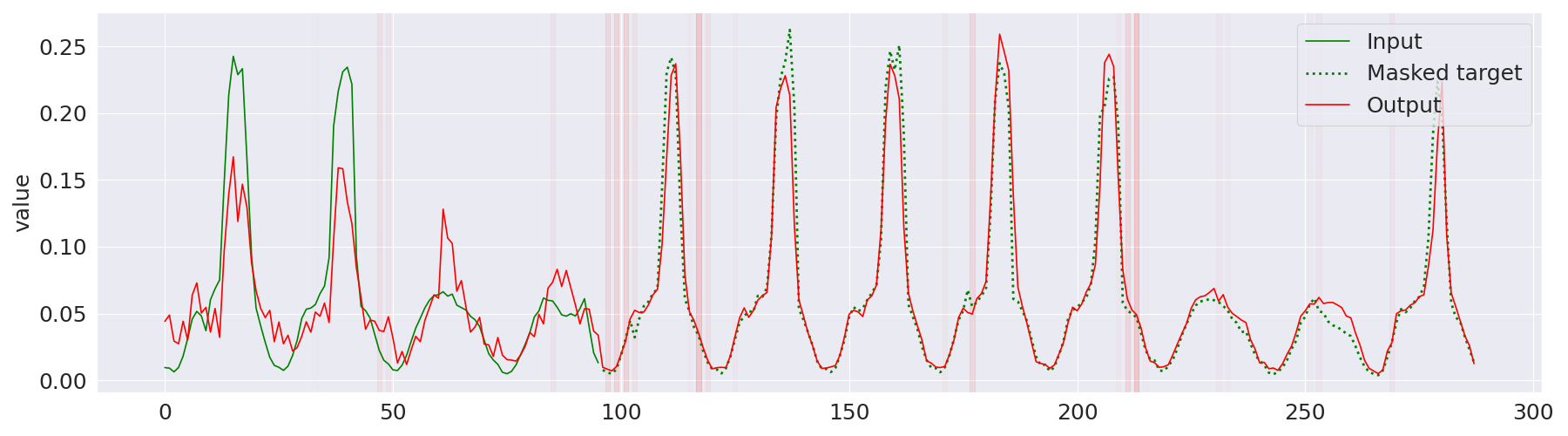}
     \end{subfigure}
     \vfill 
     \begin{subfigure}
         \centering
         \includegraphics[width=\textwidth]{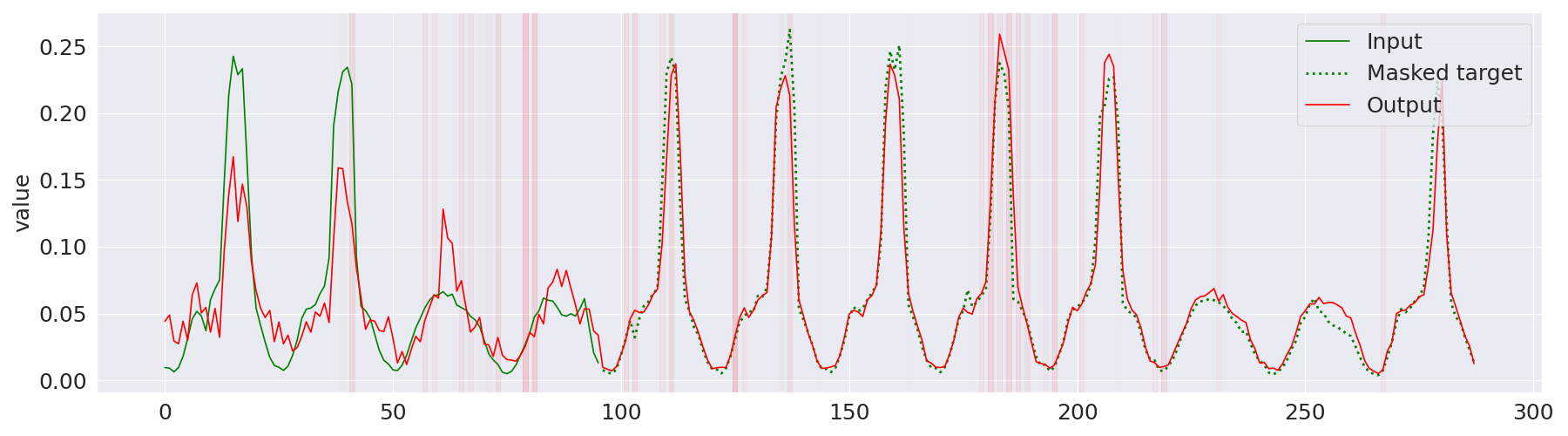}
     \end{subfigure}
     \vfill 
     \begin{subfigure}
         \centering
         \includegraphics[width=\textwidth]{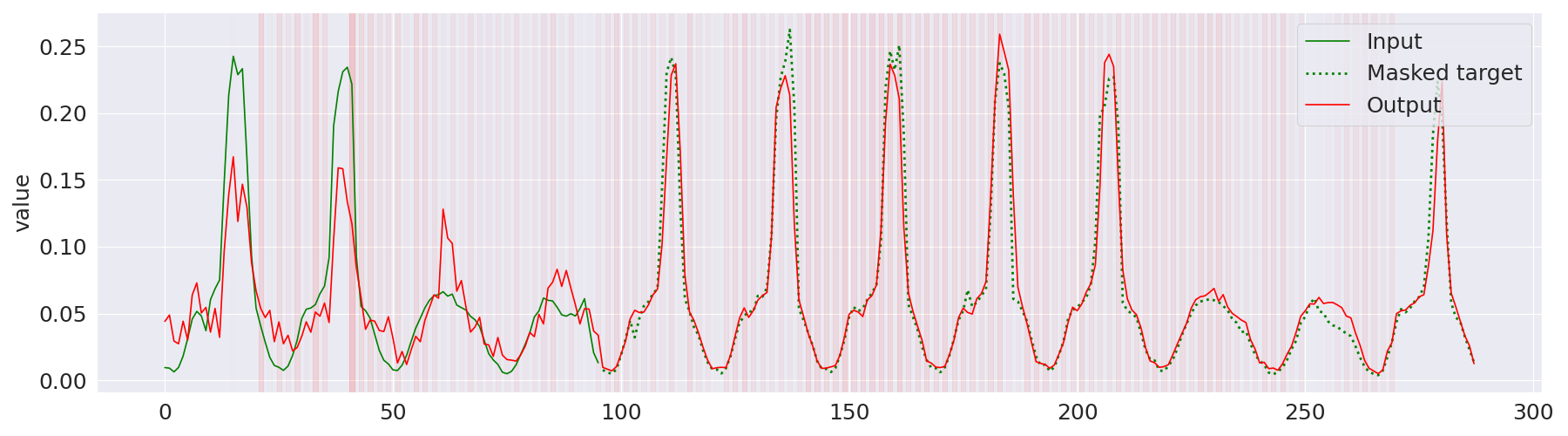}
     \end{subfigure}
     \vfill 
     \begin{subfigure}
         \centering
         \includegraphics[width=\textwidth]{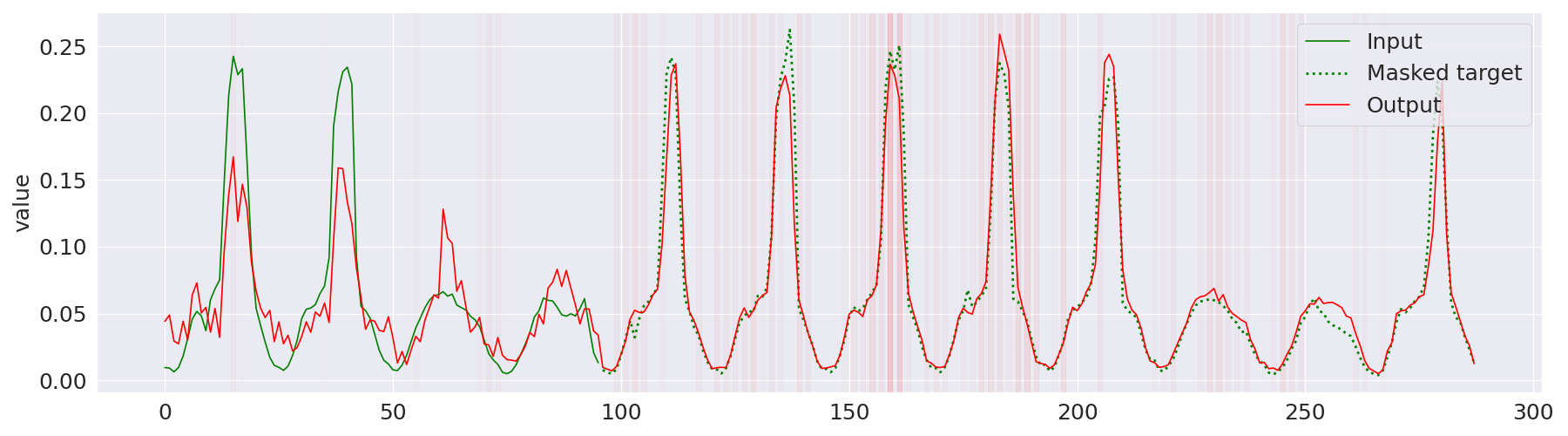}
     \end{subfigure}
        \caption{Result of a fine-tuned forecasting model on the Traffic dataset with highlighted attention weights for all 5 \EncodingLayer{}s, starting with the first \EncodingLayer{} at the top and concluding with the last at the bottom}
\end{figure}

Figure~\ref{fig:appendix:explain} shows the individual attention highlights of all 5 \EncodingLayer{}s of a fine-tuned forecasting model on the Traffic dataset as a supplement for the sum-based illustration in Section~\ref{section:explain}. The solid green line represents the initial input series, transitioning to a dotted line after the 96th value. This dotted green line delineates the target horizon, inputted into the model as $-1$ values. The order is from the top, showing the attention from the first \EncodingLayer{}, to the bottom, showing the last \EncodingLayer{}. 
We can observe that the attention from the first \EncodingLayer{} is only focused on the two last peeks before the horizon. This could indicate that the model starts the forecast by simply forward copying the last peeks, followed by mimicking the valleys, as indicated by the second \EncodingLayer{}'s attention illustrated in the respective subplot in Figure~\ref{fig:appendix:explain}. 
Later, attention seems to switch on selected representatives of repeating sub-areas in the time series (e.g. the curve rise after 75 and 100 or the peak before and the valley after 200), giving the impression that it is focusing on these areas more closely.
Remarkably, the attention of the penultimate \EncodingLayer{} tends to be highly distributed and nearly spans around the entire sequence, in contrast to the rest of the \EncodingLayer{}'s attention, which mainly focuses on a few areas. This could indicate a normalization process of the forecast before the last layer. Note that each \EncodingLayer{} has access to the result of the previous one, so attention shifts more towards the horizon with increasing progresses through the \EncodingLayer{}s. This could indicate that the model can propagate information across the \EncodingLayer{} to improve the result.

\subsection{Methodologies: Forecasting}\label{appendix:meth_forecasting}

\begin{figure}[h]
\includegraphics[width=0.8\textwidth]{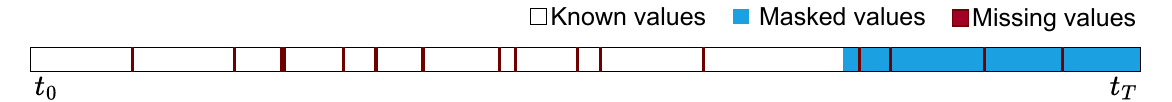}
\centering

\caption{Illustration of the setup of a forecasting task during fine-tuning.}
\label{figure:app:forecasting}
\end{figure}

In the prediction task, $-1$ tokens are used to denote the horizon and mark missing data in the input sequence. The process is illustrated in Figure~\ref{figure:app:forecasting}. The model processes only this masked sequence, with white sequences marking known values, red sequences marking missing values, and blue values marking the masked horizon. The model then predicts values corresponding to these tokens; only the values masked as the horizon are used to evaluate the effectiveness of the model.

\subsection{Experiments: Imputation}\label{appendix:exp_imputation1}
This section gives additional details about our imputation related experiments for the Electricity and the Air-Quality dataset. All experiments were part of a hyperparameter study utilizing a systematic grid search methodology. We examined the subsequent hyperparameters within these specified ranges: 
\begin{itemize}
    \item Number of \EncodingLayer{}s ($N$): 2, 5, 7, 10, 15, 20
    \item Number of heads in the self-attention module ($h$): 2, 4, 8, 16
    \item Feature embedding size ($f_{embed}$: 32, 64, 128, 256
    \item Attention function ($attention\_func$): classic (vanilla)~\cite{vaswani2017attention}, Propsparse attention\cite{zhou2021informer}, sparse-attention~\cite{wu2020adversarial}.
    \item Amount and configuration of the CNN layers in the \RepresentationLayer{}: We varied the amount of CNN layers between 1 and 4. The configuration was designed that the lowest kernel covered around 3 values and the biggest around 50\% - 80\% of the input sequence. The kernels in between covered middle sized sequences. 

\end{itemize}

\paragraph{Electricity}

In Table~\ref{tab:app:imputation_pretrain}, we outline the performance of diverse configurations of pretrained models trained on the Electricity and Air-Quality dataset for the experiments in Section~\ref{sec:meth:imputation}. The optimum performance is documented in the first line for each data set and was used for fine-tuning in the corresponding experiments. Note that the values in column "$f_{embed}$" are corresponding to the feature individual embedding size (see Section~\ref{section:model_architecture}), not the total. 

\paragraph{Air-Quality}
As outlined in Section~\ref{sec:meth:imputation}, the performance on the Air-Quality dataset did not match that of the Electricity dataset. Further analysis indicated that the shorter sequence lengths within the Air-Quality dataset limited the efficacy of our architecture to acquire meaningful representations. Additionally, the reliance of the Air-Quality dataset on temporal encodings for optimal imputation poses a challenge for our current architectural setup. Continued examination of this issue using the ETTm2 dataset is detailed in Section~\ref{appendix:exp_forecasting}.

\begin{figure}
\label{fig:appendix:imputation1}
     \centering
     \begin{subfigure}
         \centering
         \includegraphics[width=\textwidth]{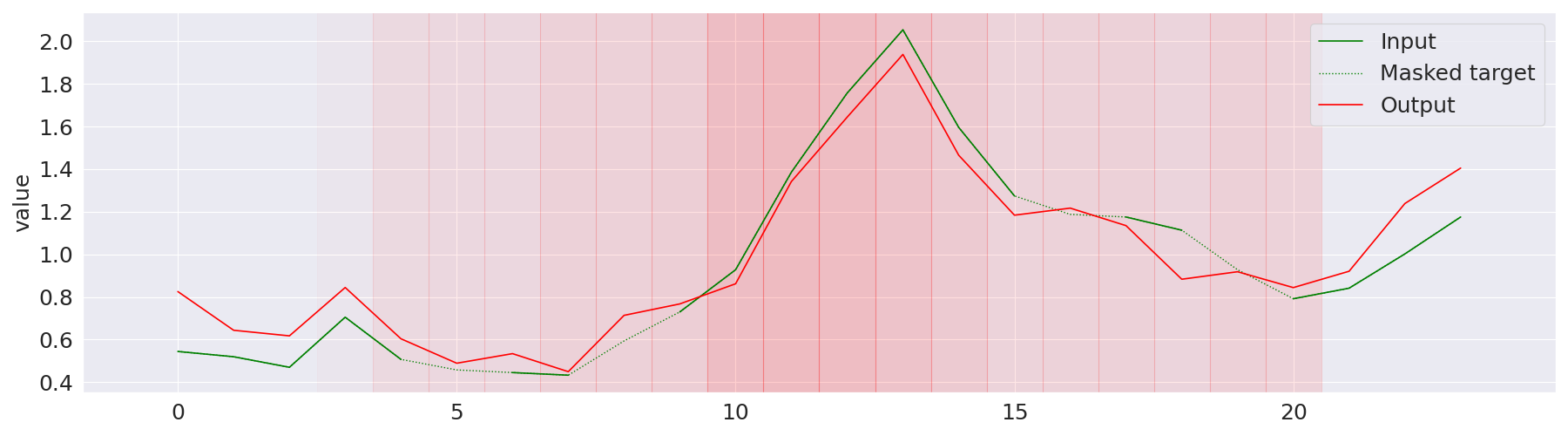}
     \end{subfigure} 
     \vfill 
     \begin{subfigure}
         \centering
         \includegraphics[width=\textwidth]{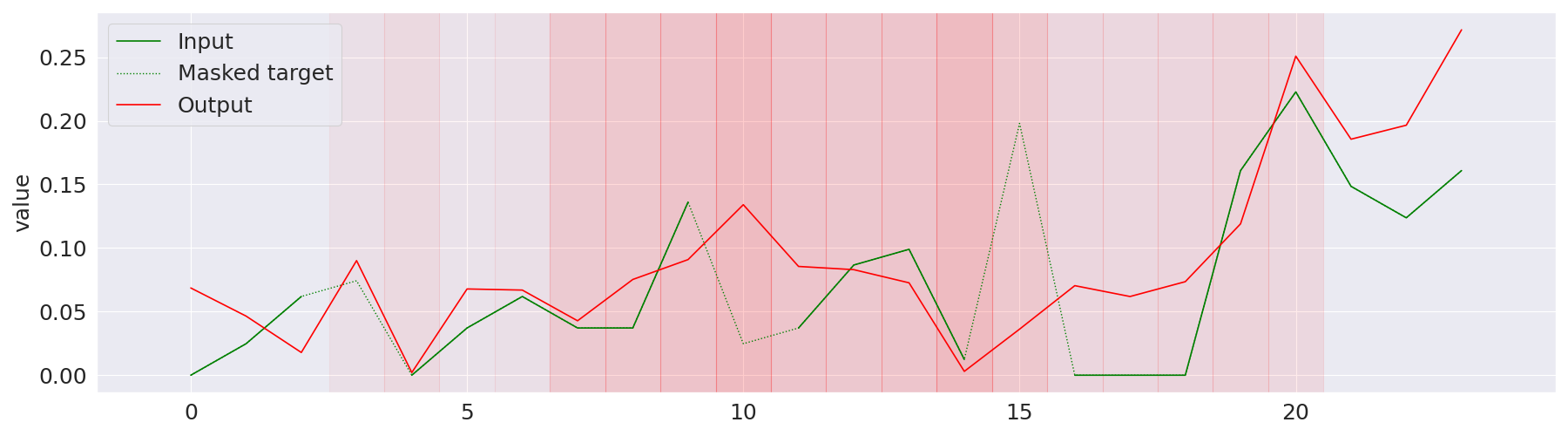}
     \end{subfigure}

        \caption{Two example results of a fine-tuned imputation model on the Air-Quality dataset with highlighted attention weights.}
\end{figure}

Figure~\ref{fig:appendix:imputation1} illustrates two exemplars of outputs from a fine-tuned imputation model applied to the Air-Quality dataset. The initial graph demonstrates the attention mechanism's ability to focus on central patterns, achieving effective resolution of the imputation challenge. In contrast, the subsequent graph reveals sequences devoid of observable patterns, necessitating reliance on temporal embeddings for quality imputation. Notably, at $t=15$, the emergent peak in missing values is unforecastable without supplementary data from temporal embeddings.

\begin{table}[]
\caption{Performance comparison of pretrained models for the imputation task on the Electricity and Air-Quality dataset.}
\centering
\label{tab:app:imputation_pretrain}
\scalebox{.9}{
\begin{tabular}{@{}ccccclccc@{}}
\hline
\multicolumn{5}{c}{Hyper-parameters} &  & \multicolumn{3}{c}{Metrics} \\ \cmidrule(r){1-5} \cmidrule(l){7-9} 
N & h & $f_{embedd}$ & attention & \begin{tabular}[c]{@{}c@{}}Representations\\ (kernel{[}\%{]}, dilation)\end{tabular} &  & \begin{tabular}[c]{@{}c@{}}Reproduce\\ {[}MSE{]}\end{tabular} & \begin{tabular}[c]{@{}c@{}}Masked $\downarrow$\\ {[}MSE{]}\end{tabular} & \begin{tabular}[c]{@{}c@{}}Classific.\\ {[}$f_1${]}\end{tabular} \\ 
\\
\hline
\multicolumn{9}{c}{Electricity} \\ \hline

\rowcolor[HTML]{C0C0C0} 3 & 4 & 128 & classic &  \begin{tabular}[c]{@{}l@{}}k: 3, d: 2 \end{tabular} & & 0.001 & 0.04 & 1.0 \\ 5 & 2 & 128 & entmax15 &  \begin{tabular}[c]{@{}l@{}}k: 3, d: 2 \end{tabular} & & 0.002 & 0.041 & 1.0 \\ \rowcolor[HTML]{C0C0C0} 3 & 2 & 128 & classic &  \begin{tabular}[c]{@{}l@{}}k: 3, d: 2 \\ k: 10, d: 4 \\ k: 30, d: 4 \end{tabular} & & 0.003 & 0.042 & 1.0 \\ 5 & 8 & 128 & entmax15 &  \begin{tabular}[c]{@{}l@{}}k: 3, d: 2 \end{tabular} & & 0.005 & 0.043 & 1.0 \\ \rowcolor[HTML]{C0C0C0} 5 & 2 & 64 & entmax15 &  \begin{tabular}[c]{@{}l@{}}k: 3, d: 2 \end{tabular} & & 0.009 & 0.053 & 1.0 \\ 10 & 2 & 128 & propsparse &  \begin{tabular}[c]{@{}l@{}}k: 3, d: 2 \\ k: 10, d: 4 \\ k: 60, d: 4 \end{tabular} & & 0.022 & 0.058 & 1.0 \\ \rowcolor[HTML]{C0C0C0} 3 & 4 & 32 & entmax15 &  \begin{tabular}[c]{@{}l@{}}k: 3, d: 4 \\ k: 30, d: 4 \end{tabular} & & 0.017 & 0.061 & 0.996 \\ 3 & 2 & 32 & classic &  \begin{tabular}[c]{@{}l@{}}k: 3, d: 2 \\ k: 10, d: 4 \\ k: 60, d: 4 \end{tabular} & & 0.033 & 0.065 & 0.889 \\  

\\
\hline
\multicolumn{9}{c}{Air-Quality} \\ \hline
\rowcolor[HTML]{C0C0C0} 5 & 4 & 32 & propsparse &  \begin{tabular}[c]{@{}l@{}}k: 5, d: 1 \\ k: 20, d: 1 \end{tabular} & & 0.038 & 0.149 & 1.0 \\ 3 & 2 & 16 & propsparse &  \begin{tabular}[c]{@{}l@{}}k: 5, d: 1 \\ k: 20, d: 1 \end{tabular} & & 0.046 & 0.151 & 1.0 \\ \rowcolor[HTML]{C0C0C0} 10 & 8 & 32 & classic &  \begin{tabular}[c]{@{}l@{}}k: 10, d: 1 \\ k: 20, d: 1 \\ k: 60, d: 1 \end{tabular} & & 0.047 & 0.153 & 1.0 \\ 5 & 8 & 16 & entmax15 &  \begin{tabular}[c]{@{}l@{}}k: 10, d: 1 \\ k: 30, d: 1 \end{tabular} & & 0.052 & 0.153 & 1.0 \\ 

\bottomrule
\end{tabular}
}
\end{table}

In addition to the results presented in Section~\ref{sec:meth:imputation} we report in Table~\ref{tab:app:imputation1} the performance of our fine-tuned imputation model in the Electricity dataset with 30\%, 50\%, 70\%, and 90\% data randomly omitted. 

\begin{table}[]
\caption{Performance comparison for the imputation task on the Electricity dataset.}
\centering
\label{tab:app:imputation1}
\scalebox{.9}{
\begin{tabular}{@{}llllllllll@{}}
\toprule
Missing & \multicolumn{2}{c}{30\%} & \multicolumn{2}{c}{50\%} & \multicolumn{2}{c}{70\%} & \multicolumn{2}{c}{90\%} &  \\ \midrule
Metric & {\color[HTML]{232627} MAE} & {\color[HTML]{232627} RMSE} & {\color[HTML]{232627} MAE} & {\color[HTML]{232627} RMSE} & {\color[HTML]{232627} MAE} & {\color[HTML]{232627} RMSE} & MAE & RMSE $\downarrow$ & P{[}M{]} \\ \midrule
M-RNN & 1.258 & 1.876 & 1.283 & 1.902 & 1.305 & 1.928 & 1.331 & 1.961 & 18.6 \\
BRITS & 0.943 & 1.435 & 1.037 & 1.538 & 1.090 & 1.617 & 1.163 & 1.702 & 7.00 \\
GP-VAE & 1.057 & 1.571 & 1.097 & 1.572 & 1.037 & 1.598 & 1.004 & 1.622 & 13.5 \\
Transf. & 0.846 & 1.321 & 0.895 & 1.410 & 0.920 & 1.437 & 0.934 & 1.492 & 14.8 \\
SAITS & 0.790 & 1.223 & 0.876 & 1.377 & 0.898 & 1.273 & 0.933 & 1.354 & 11.5 \\
TSRM & \textbf{0.075} & \textbf{0.156} & \textbf{0.089} & \textbf{0.177} & \textbf{0.139} & \textbf{0.278} & \textbf{0.273} & \textbf{0.506} & \textbf{0.86} \\ \bottomrule
\end{tabular}
}
\end{table}

\paragraph{Multiple random runs}
To evaluate the significance of our imputation experiments we provide the means and standard deviation (STD) for both datasets in Table~\ref{appendix:tab:imputation_signi}. To calculate the mean and the STD we performed all runs five times with different seeds. 

\begin{table}[]
\caption{Imputation benchmarks based on five runs with different seeds.}
\centering
\label{appendix:tab:imputation_signi}
\begin{tabular}{@{}lll@{}}
\toprule
Dataset & MSE & MAE \\ \midrule
Electricity & 0.023 $\pm$ 0.0004 & 0.077 $\pm$ 0.0056 \\
Air-Quality & 0.140 $\pm$ 0.0191 & 0.169 $\pm$ 0.0054 \\ \bottomrule
\end{tabular}
\end{table}

\subsection{Experiments: Forecasting}\label{appendix:exp_forecasting}

Table~\ref{tab:app:imputation_forecating1} shows the performance of diverse configurations of pretrained models evaluated on the Electricity dataset, with the best performing configuration listed in the first row.
Note, that all experiments and hyperparamter studies were equally designed as described in Section~\ref{appendix:exp_imputation1}.

\begin{table}[]
\caption{Performance comparison of pretrained models for the imputation task on the Air-Quality dataset.}
\centering
\label{tab:app:imputation_forecating1}
\scalebox{.9}{
\begin{tabular}{@{}ccccclccc@{}}
\toprule
\multicolumn{5}{c}{Hyper-parameters} &  & \multicolumn{3}{c}{Metrics} \\ \cmidrule(r){1-5} \cmidrule(l){7-9} 
N & h & $d_{embedd}$ & attention & \begin{tabular}[c]{@{}c@{}}Representations\\ (kernel{[}\%{]}, dilation)\end{tabular} &  & \begin{tabular}[c]{@{}c@{}}Reproduce\\ {[}MSE{]}\end{tabular} & \begin{tabular}[c]{@{}c@{}}Masked $\downarrow$\\ {[}MSE{]}\end{tabular} & \begin{tabular}[c]{@{}c@{}}Classific.\\ {[}$f_1${]}\end{tabular} \\
\rowcolor[HTML]{C0C0C0} 5 & 4 & 32 & propsparse &  \begin{tabular}[c]{@{}l@{}}k: 5, d: 1 \\ k: 20, d: 1 \end{tabular} & & 0.038 & 0.149 & 1.0 \\ 3 & 2 & 16 & propsparse &  \begin{tabular}[c]{@{}l@{}}k: 5, d: 1 \\ k: 20, d: 1 \end{tabular} & & 0.046 & 0.151 & 1.0 \\ \rowcolor[HTML]{C0C0C0} 10 & 8 & 32 & classic &  \begin{tabular}[c]{@{}l@{}}k: 10, d: 1 \\ k: 20, d: 1 \\ k: 60, d: 1 \end{tabular} & & 0.047 & 0.153 & 1.0 \\ 5 & 8 & 16 & entmax15 &  \begin{tabular}[c]{@{}l@{}}k: 10, d: 1 \\ k: 30, d: 1 \end{tabular} & & 0.052 & 0.153 & 1.0 \\   \bottomrule
\end{tabular}
}
\end{table}

\begin{table}[]
\caption{Performance comparison of pretrained models for the forecasting task on the Electricity, Traffic and ETTm2 dataset.}
\centering
\label{tab:app:imputation_pretrain}
\scalebox{.8}{
\begin{tabular}{@{}ccccclccc@{}}
\hline
\multicolumn{5}{c}{Hyper-parameters} &  & \multicolumn{3}{c}{Metrics} \\ \cmidrule(r){1-5} \cmidrule(l){7-9} 
N & h & $d_{embedd}$ & attention & \begin{tabular}[c]{@{}c@{}}Representations\\ (kernel{[}\%{]}, dilation)\end{tabular} &  & \begin{tabular}[c]{@{}c@{}}Reproduce\\ {[}MSE{]}\end{tabular} & \begin{tabular}[c]{@{}c@{}}Masked $\downarrow$\\ {[}MSE{]}\end{tabular} & \begin{tabular}[c]{@{}c@{}}Classific.\\ {[}$f_1${]}\end{tabular} \\ 
\\
\hline
\multicolumn{9}{c}{Electricity} \\ \hline

\rowcolor[HTML]{C0C0C0} 3 & 2 & 256 & classic &  \begin{tabular}[c]{@{}l@{}}k: 1, d: 2 \\ k: 10, d: 4 \\ k: 60, d: 4 \end{tabular} & & 0.0 & 0.031 & 1.0 \\ 3 & 2 & 256 & classic &  \begin{tabular}[c]{@{}l@{}}k: 1, d: 2 \\ k: 10, d: 4 \\ k: 60, d: 4 \end{tabular} & & 0.0 & 0.034 & 1.0 \\ \rowcolor[HTML]{C0C0C0} 3 & 2 & 256 & classic &  \begin{tabular}[c]{@{}l@{}}k: 1, d: 2 \\ k: 10, d: 4 \\ k: 60, d: 4 \end{tabular} & & 0.001 & 0.038 & 0.886 \\ 3 & 2 & 128 & classic &  \begin{tabular}[c]{@{}l@{}}k: 1, d: 2 \\ k: 10, d: 4 \\ k: 60, d: 4 \end{tabular} & & 0.0 & 0.038 & 0.886 \\ \rowcolor[HTML]{C0C0C0} 10 & 4 & 128 & classic &  \begin{tabular}[c]{@{}l@{}}k: 5, d: 4 \\ k: 30, d: 4 \end{tabular} & & 0.024 & 0.066 & 1.0 \\   

\\
\hline
\multicolumn{9}{c}{Air-Quality} \\ \hline
\rowcolor[HTML]{C0C0C0} 5 & 2 & 128 & propsparse &  \begin{tabular}[c]{@{}l@{}}k: 1, d: 2 \\ k: 10, d: 4 \\ k: 60, d: 4 \end{tabular} & & 0.0 & 0.001 & 1.0 \\ 5 & 2 & 64 & entmax15 &  \begin{tabular}[c]{@{}l@{}}k: 1, d: 2 \\ k: 10, d: 4 \\ k: 60, d: 4 \end{tabular} & & 0.0 & 0.001 & 1.0 \\ \rowcolor[HTML]{C0C0C0} 5 & 4 & 128 & classic &  \begin{tabular}[c]{@{}l@{}}k: 1, d: 2 \\ k: 10, d: 4 \\ k: 60, d: 4 \end{tabular} & & 0.0 & 0.001 & 1.0 \\ 5 & 8 & 128 & propsparse &  \begin{tabular}[c]{@{}l@{}}k: 1, d: 2 \\ k: 10, d: 4 \\ k: 60, d: 4 \end{tabular} & & 0.0 & 0.001 & 1.0 \\ \rowcolor[HTML]{C0C0C0} 5 & 4 & 64 & classic &  \begin{tabular}[c]{@{}l@{}}k: 1, d: 2 \\ k: 10, d: 4 \\ k: 60, d: 4 \end{tabular} & & 0.0 & 0.001 & 0.889 \\ 

\\
\hline
\multicolumn{9}{c}{ETTm2} \\ \hline

\rowcolor[HTML]{C0C0C0} 3 & 4 & 128 & entmax15 &  \begin{tabular}[c]{@{}l@{}}k: 1, d: 1 \\ k: 5, d: 2 \\ k: 20, d: 4 \end{tabular} & & 0.001 & 0.005 & 0.894 \\ 5 & 2 & 128 & classic &  \begin{tabular}[c]{@{}l@{}}k: 1, d: 1 \\ k: 5, d: 2 \\ k: 20, d: 4 \end{tabular} & & 0.001 & 0.005 & 0.888 \\ \rowcolor[HTML]{C0C0C0} 3 & 4 & 256 & propsparse &  \begin{tabular}[c]{@{}l@{}}k: 1, d: 1 \\ k: 20, d: 2 \\ k: 80, d: 4 \end{tabular} & & 0.001 & 0.005 & 0.863 \\ 3 & 8 & 256 & classic &  \begin{tabular}[c]{@{}l@{}}k: 1, d: 1 \\ k: 5, d: 2 \\ k: 20, d: 4 \end{tabular} & & 0.001 & 0.006 & 0.0 \\ \rowcolor[HTML]{C0C0C0} 5 & 8 & 128 & propsparse &  \begin{tabular}[c]{@{}l@{}}k: 1, d: 1 \\ k: 20, d: 2 \\ k: 80, d: 4 \end{tabular} & & 0.001 & 0.006 & 0.918 \\ 

\bottomrule
\end{tabular}
}
\end{table}

\begin{figure}
\label{fig:appendix:forecasting}
     \centering
     \begin{subfigure}
         \centering
         \includegraphics[width=\textwidth]{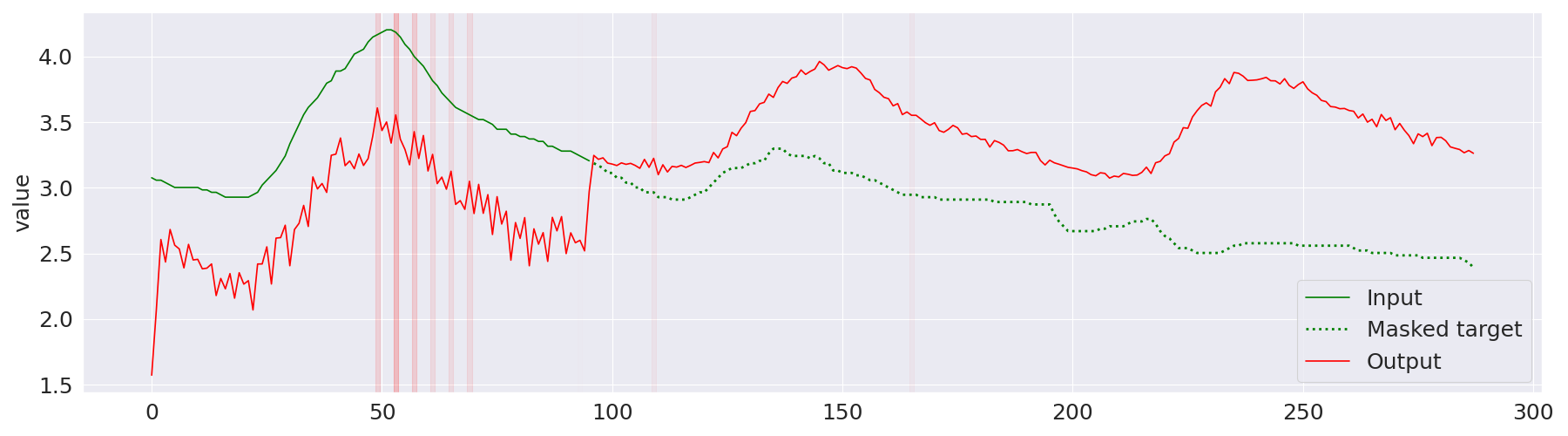}
     \end{subfigure} 
     \vfill 
     \begin{subfigure}
         \centering
         \includegraphics[width=\textwidth]{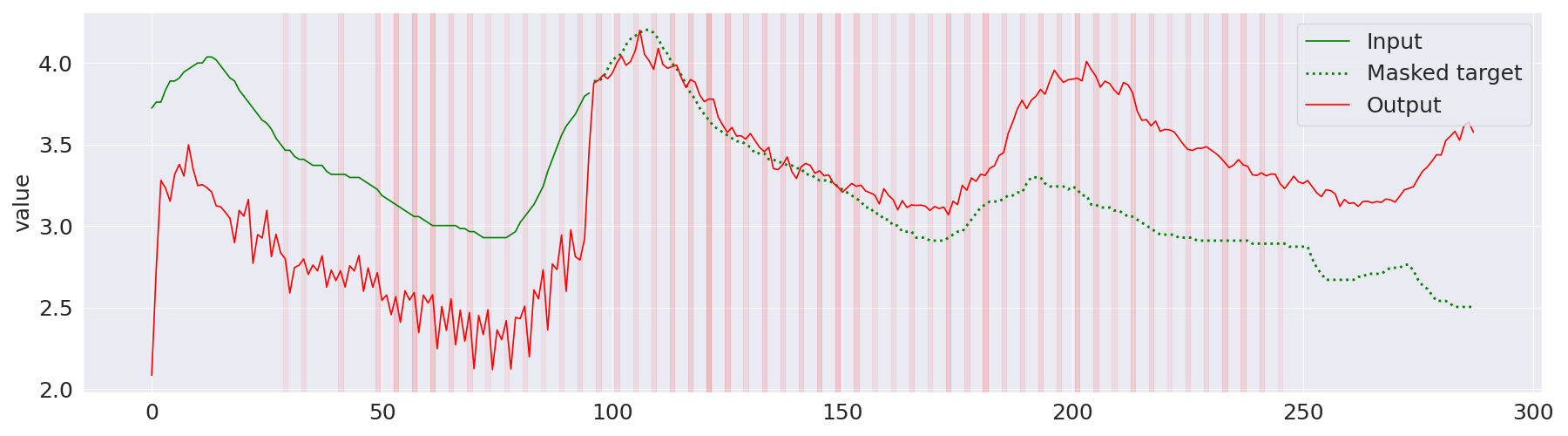}
     \end{subfigure}

        \caption{Two example results of a fine-tuned forecasting model on the ETTm2 dataset with highlighted attention weights.}
\end{figure}

As shown in the main experiments in Section~\ref{sec:exp:forecasting} our model performed considerably well on the Electricity and the Traffic datasets. On the ETTm2, however, we could hardly match SOTA performance. Further investigation revealed that temporal embedding, employed by listed baseline methods, is a crucial part of a successful forecast since the univariate input time series does not provide sufficient information for high-quality forecasting. Despite experiments involving equal temporal embeddings, we could not improve the results on the ETTm2 dataset. This indicates that the architecture struggles to extract and utilize information from embeddings, such as temporal and positional, for its task.
Figure~\ref{fig:appendix:forecasting} illustrates two forecasting scenarios using the ETTm2 dataset, where the input sequence is depicted as a solid green line and the predictive horizon as a dotted green line. The output generated by our model is represented by a red line, with attention summed up across all \EncodingLayer{} layers. In the initial graph, attention is primarily directed towards the last peak of the input sequence, attempting to replicate this pattern in the forecasted data. Nonetheless, the series subsequently exhibits a declining trend devoid of identifiable structures. This pattern suggests potential temporal correlations related to specific time variables, such as hours or months, embedded within the time series. Given that our architecture cannot accurately forecast time series with embedded temporal markers, we infer a limitation in our architecture's ability to extract information from embeddings adequately.
The secondary graph in Figure~\ref{fig:appendix:forecasting} depicts a scenario wherein the model attempts to replicate the established structural patterns from the input, yet the trajectory of the horizon transitions to a declining trend. In this case, the attention is distributed throughout the time series at defined intervals, potentially correlating with the temporal embedding applied.
We will investigate this issues regarding temporal and positional embeddings in future work.

In the context of trainable parameters, our findings indicate a substantial reduction in trainable parameters required for optimal model performance across various datasets. Specifically, for the ETTm2 dataset, the best model necessitates 3.4M parameters, while the Electricity and Traffic datasets require 6.3M and 10.4M parameters, respectively. This compares favorably to state-of-the-art methods, where leading models such as FEDformer, Informer, and Autoformer each demand 16.3M trainable parameters.

\paragraph{Multiple random runs}
To evaluate the significance of our forecasting experiments we provide the means and standard deviation (STD) for all three datasets in Table~\ref{appendix:tab:forecast_signi}. To calculate the mean and the STD we performed all runs five times with different seeds. 

\begin{table}[]
\caption{Forecasting benchmarks based on five runs with different seeds.}
\centering
\label{appendix:tab:forecast_signi}
\begin{tabular}{@{}lll@{}}
\toprule
Dataset & MSE & MAE \\ \midrule
Electricity & 0.081 $\pm$ 0.74e-3 & 0.129 $\pm$ 0.48e-2 \\
Traffic & 0.001 $\pm$ 0.22e-4 & 0.0181 $\pm$ 0.67e-4 \\
ETTm2 & 0.125 $\pm$ 0.002 & 0.271 $\pm$ 0.01 \\\bottomrule
\end{tabular}
\end{table}

\subsection{Experiments: Classification}\label{appendix:meth_classification}
This section gives an additional evaluation of our architecture regarind time series classification.
Note, that all experiments and hyperparamter studies were equally designed as described in Section~\ref{appendix:exp_imputation1}.

The procedure of time series classification involves utilizing the pre-trained \AttMapClassifier{}. 
During the fine-tuning process, the final three linear components of the \AttMapClassifier{} are substituted with untrained equivalents tailored to the requisite class outputs. Concurrently, modifications are made to the loss function's weights to exclusively emphasize the classification loss $L_{Class.}$. As \EncodingLayer{}s represent the majority of trainable parameters and remain inactive in classification fine-tuning, the adjustment scope for a TSRM in this context is confined to approximately ten thousand trainable parameters of the \AttMapClassifier{}.

In order to assess the efficacy of our architecture in the context of time series classification, we conduct model training utilizing the WISDM dataset. We perform a comparative analysis against contemporary methodologies: Park et al.~\cite{park2023multicnn}, Zhang et al.~\cite{zhang2019novel}, and Singh et al.~\cite{singh2020deep}.
We stick to established benchmark to enhance comparability and adopt the setup prescribed by Park et al.~\cite{park2023multicnn}.

A TSRM was pre-trained on the WISDM dataset, and the optimal variant was selected for enhanced optimization and targeted at the classification task.  Despite numerous experimental iterations, we were unable to achieve the current SOTA results demonstrated in studies by Park et al.~\cite{park2023multicnn} ($F_1$: 92.50), Zhang et al.~\cite{zhang2019novel} ($F_1$: 90.42), and Singh et al.~\cite{singh2020deep} ($F_1$: 89.61). Our most effective model achieved a $F_1$ score of 0.81, indicating the necessity for further refinements in time series classification methodologies.

Table~\ref{tab:app:classification_pretrain} presents the efficacy of various pretrained model configurations on the WISDM dataset, with the top-performing setup detailed at the top. Across all configurations, classification outcomes were suboptimal, rarely achieving an $f_1$ score exceeding 0.88. This suggests that the models could not derive adequate representations from the dataset, impeding their ability to address subsequent fine-tuning tasks effectively.

\begin{table}[]
\caption{Performance comparison of pretrained models for the classification task on the WISDM dataset.}
\centering
\label{tab:app:classification_pretrain}
\scalebox{.8}{
\begin{tabular}{@{}ccccclccc@{}}
\hline
\multicolumn{5}{c}{Hyper-parameters} &  & \multicolumn{3}{c}{Metrics} \\ \cmidrule(r){1-5} \cmidrule(l){7-9} 
N & h & $d_{embedd}$ & attention & \begin{tabular}[c]{@{}c@{}}Representations\\ (kernel{[}\%{]}, dilation)\end{tabular} &  & \begin{tabular}[c]{@{}c@{}}Reproduce\\ {[}MSE{]}\end{tabular} & \begin{tabular}[c]{@{}c@{}}Masked $\downarrow$\\ {[}MSE{]}\end{tabular} & \begin{tabular}[c]{@{}c@{}}Classific.\\ {[}$f_1${]}\end{tabular} \\ 

\rowcolor[HTML]{C0C0C0} 3 & 2 & 128 & entmax15 &  \begin{tabular}[c]{@{}l@{}}k: 4, d: 1 \\ k: 10, d: 2 \\ k: 30, d: 3 \end{tabular} & & 0.015 & 0.296 & 0.884 \\ 5 & 2 & 128 & propsparse &  \begin{tabular}[c]{@{}l@{}}k: 4, d: 1 \\ k: 10, d: 2 \\ k: 30, d: 3 \end{tabular} & & 0.02 & 0.31 & 0.868 \\ \rowcolor[HTML]{C0C0C0} 5 & 2 & 128 & propsparse &  \begin{tabular}[c]{@{}l@{}}k: 4, d: 1 \\ k: 20, d: 2 \\ k: 40, d: 3 \end{tabular} & & 0.015 & 0.318 & 0.892 \\ 7 & 4 & 128 & propsparse &  \begin{tabular}[c]{@{}l@{}}k: 10, d: 1 \\ k: 30, d: 2 \\ k: 50, d: 3 \end{tabular} & & 0.037 & 0.339 & 0.885 \\ 

\\

\bottomrule
\end{tabular}
}
\end{table}

\end{document}